\documentclass[10pt,twocolumn,letterpaper]{article}

\usepackage{cvpr}              

\usepackage[dvipsnames]{xcolor}
\usepackage{tabularx}
\usepackage{boldline}

\usepackage[normalem]{ulem}

\DeclareMathOperator*{\argmin}{argmin}

\definecolor{cvprblue}{rgb}{0.21,0.49,0.74}
\usepackage[pagebackref,breaklinks,colorlinks,allcolors=cvprblue]{hyperref}

\def\paperID{5875} 
\def\confName{CVPR}
\def\confYear{2025}

\title{Ego4o: Egocentric Human Motion Capture and Understanding from Multi-Modal Input}

\author{Jian Wang\textsuperscript{1,4}~~~~~~Rishabh Dabral\textsuperscript{1,4}~~~~~~Diogo Luvizon\textsuperscript{1,4}~~~~~~ Zhe Cao\textsuperscript{2}\\~~~~~~Lingjie Liu\textsuperscript{3}~~~~~~Thabo Beeler\textsuperscript{2}~~~~~~Christian Theobalt\textsuperscript{1,4}\\
\textsuperscript{1}MPI Informatics \& Saarland Informatics Campus~~~~~\textsuperscript{2}Google~~~~~\textsuperscript{3}University of Pennsylvania \\ \textsuperscript{4}Saarbrücken Research Center for Visual Computing, Interaction and Artificial Intelligence
}

\begin{document}
\twocolumn[{
\maketitle
\begin{center}
    \captionsetup{type=figure}
    \includegraphics[width=1\textwidth]{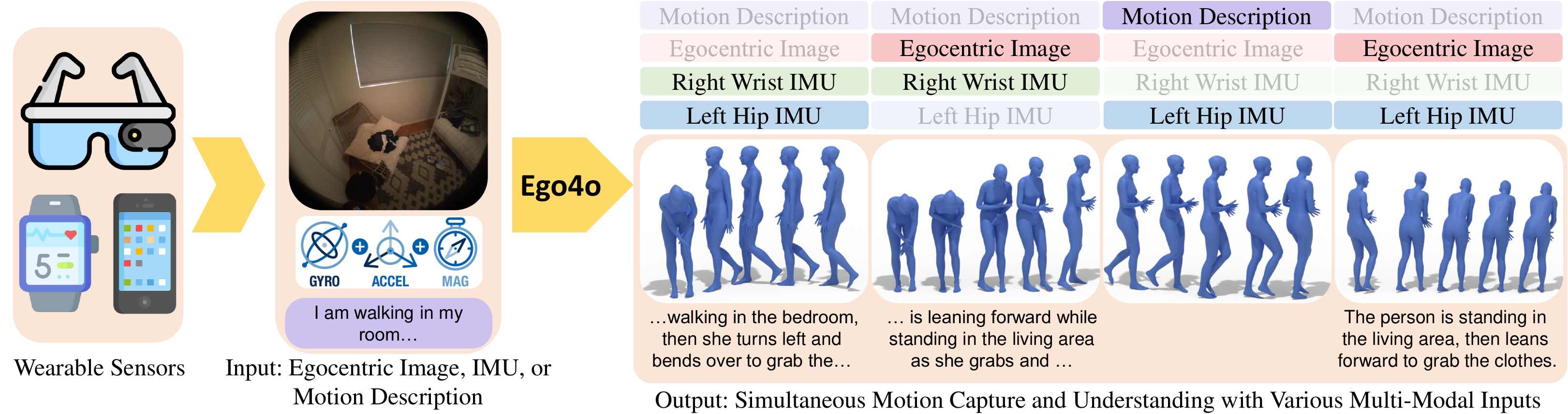}
    \captionof{figure}{
Our method can use an egocentric image and 1-3 IMU sensors from wearable devices to accurately predict human motion and generate motion descriptions. Motion descriptions, when available, can also enhance motion capture accuracy. Ego4o supports flexible input combinations, functioning with or without images, or with varied IMU placements.
}
\label{fig:teaser}
\end{center}
}]

\begin{abstract}
This work focuses on tracking and understanding human motion using consumer wearable devices, such as VR/AR headsets, smart glasses, cellphones, and smartwatches. These devices provide diverse, multi-modal sensor inputs, including egocentric images, and 1-3 sparse IMU sensors in varied combinations. Motion descriptions can also accompany these signals. The diverse input modalities and their intermittent availability pose challenges for consistent motion capture and understanding. In this work, we present \textit{Ego4o} (\textit{o} for omni), a new framework for simultaneous human motion capture and understanding from multi-modal egocentric inputs. This method maintains performance with partial inputs while achieving better results when multiple modalities are combined. First, the IMU sensor inputs, the optional egocentric image, and text description of human motion are encoded into the latent space of a motion VQ-VAE. Next, the latent vectors are sent to the VQ-VAE decoder and optimized to track human motion. When motion descriptions are unavailable, the latent vectors can be input into a multi-modal LLM to generate human motion descriptions, which can further enhance motion capture accuracy. Quantitative and qualitative evaluations demonstrate the effectiveness of our method in predicting accurate human motion and high-quality motion descriptions. Project page: \href{https://jianwang-mpi.github.io/ego4o/}{https://jianwang-mpi.github.io/ego4o}.
\vspace{-1em}
\end{abstract}    
\section{Introduction}\label{sec:intro}
Recently, more and more research has focused on human motion capture and understanding using widely available wearable devices, such as VR/AR headsets, smart glasses, cellphones, and smartwatches~\cite{luo2021dynamics, li2023ego, wang2024egocentric, ma2024nymeria, mollyn2023imuposer, van2024diffusionposer, xu2024mobileposer}. 
This interest is driven by broad application scenarios, including sports, healthcare, VR/AR, and personal assistants.
These devices provide diverse, multi-modal sensor inputs related to human motion, such as inertial measurement unit (IMU) data, egocentric camera images, and even voice-enabled conversation data, where the text description of human motion can be extracted. 
However, existing works mostly focus on motion capture from one single input modality. Some methods~\cite{luo2021dynamics, li2023ego, wang2024egocentric} predict the human motion from egocentric cameras, while others capture the human motion from VR tracker~\cite{jiang2022avatarposer, castillo2023bodiffusion} or IMU signals~\cite{mollyn2023imuposer, yi2022physical}. 
Each individual modality provides only a limited view of human motion, constraining the accuracy of both motion capture and understanding. For example, the full human body motion is barely seen or significantly occluded from the egocentric camera. 
Text descriptions can offer information about human motion categories but lack the precision to detail specific movements. 
IMUs on VR/AR headsets, smartwatches, and cellphones are very sparse: they usually only track the movement of one or two limbs, as people rarely wear watches on both wrists. Moreover, IMU-based methods struggle with static pose estimation due to the absence of dynamic acceleration signals.

We observe that different input modalities serve complementary roles in motion analysis. Motion descriptions and egocentric images provide rich semantic context about both the activity being performed and the environmental setting. For example, when the egocentric view shows a desk in close proximity, it strongly indicates that the person is sitting. IMU signals from wearable devices capture precise kinematic data for specific body segments. For instance, a smartwatch's IMU sensors can capture detailed hand movements, enabling the system to differentiate between blocking and smashing motions in table tennis, which may be indiscernible from the egocentric camera perspective alone. 

To fully leverage the information from wearable devices, we present \textbf{Ego4o} (o for omni), a novel framework that achieves 3D pose estimation (motion capture) and motion description generation (motion understanding) by fusing multi-modal inputs. These inputs may include 1–3 sparse IMU sensors, egocentric images, and motion descriptions from everyday wearable devices. As illustrated in \cref{fig:teaser}, the Ego4o maintains robust performance with different combinations of input modalities.

The Ego4o method first employs a multi-modal transformer to encode diverse inputs into motion codes in a part-based discrete motion representation space, which is constructed with a VQ-VAE~\cite{van2017neural}. 
Since input availability can change during use—as users may disable egocentric cameras and microphones, or vary the number of IMU sensors by removing phones or wearing smartwatches—we implement a random masking strategy during training. This approach enables the model to adapt seamlessly to different combinations of input modalities.
Finally, the motion codes are decoded into human motion predictions and refined with test-time optimization in the VQ-VAE's latent space.

Building on the previous step, we demonstrate that the obtained motion codes can be utilized as input for Large Language Models (LLMs) to generate detailed descriptions of human movement. We developed a multi-modal joint training approach that fine-tunes LLMs to simultaneously process both motion codes and egocentric images. Our work shows that LLMs' inherent strengths in in-context reasoning and image understanding can be effectively leveraged to generate high-quality motion descriptions.

Obtaining human-produced motion descriptions is typically challenging in real-world scenarios, which can limit the motion capture performance. Our insight is that the generated high-quality motion descriptions can serve as valuable conditioning signals that enhance the accuracy of our motion capture system when the human-provided motion description is absent. This introduces a feedback loop between motion capture and understanding, advancing the state-of-the-art in both tasks.

We validate Ego4o's effectiveness through quantitative and qualitative evaluations. Experimental results demonstrate that our proposed method achieves better motion capture accuracy, while simultaneously generating detailed descriptions of human movements. By integrating motion capture and motion description generation within a unified framework, Ego4o advances toward making motion analysis accessible and practical for everyday applications with consumer devices. In summary, our contributions are:

\begin{itemize}
    \item We introduce Ego4o, a novel framework that flexibly integrates multi-modal egocentric inputs to enable simultaneous motion capture and description generation;
    \item  We design a multi-modal encoder with a random masking training strategy to accommodate varying combinations of input modalities;
    \item We employ multi-modal joint fine-tuning of large language models to bridge modality gaps and support accurate motion description generation;
    \item We show that AI-generated motion descriptions can improve the accuracy of egocentric human motion capture.
\end{itemize}

\section{Related Work}
\label{sec:related_work}

\noindent\textbf{Egocentric Human Motion Capture.}
Recently, there has been growing interest in estimating egocentric 3D poses from body-worn devices. Some methods~\cite{li2023ego, jiang2017seeing, ng2020you2me, yuan20183d, yuan2019ego, luo2021dynamics} leverage head-mounted, front-facing cameras to infer motion from head movements, while others~\cite{jiang2022avatarposer, winkler2022questsim, lee2023questenvsim, du2023avatars, jiang2023egoposer, yang2024divatrack} employ three-point trackers for motion capture. Additional approaches~\cite{xu2019mo, tome2019xr, park2023domain, liu2022ego+, liu2023egofish3d, wang2021estimating, akada2022unrealego, wang2023scene, wang2022estimating, wang2024egocentric, akada20243d} use down-facing fisheye cameras to capture full-body movement, while others~\cite{guzov2021human, guzov2024interaction, jiang2022transformer, yi2022physical} rely on IMUs for body tracking. Similar to our method, recent works such as IMUPoser~\cite{mollyn2023imuposer}, MobilePoser~\cite{xu2024mobileposer}, and Diffusion-Poser~\cite{van2024diffusionposer} use 1-3 IMU sensors to capture human motion. While these methods have significantly advanced the field, they primarily focus on single-modality solutions, leaving the potential of multi-modal integration largely unexplored.

\begin{figure*}
\centering
\includegraphics[width=0.97\linewidth]{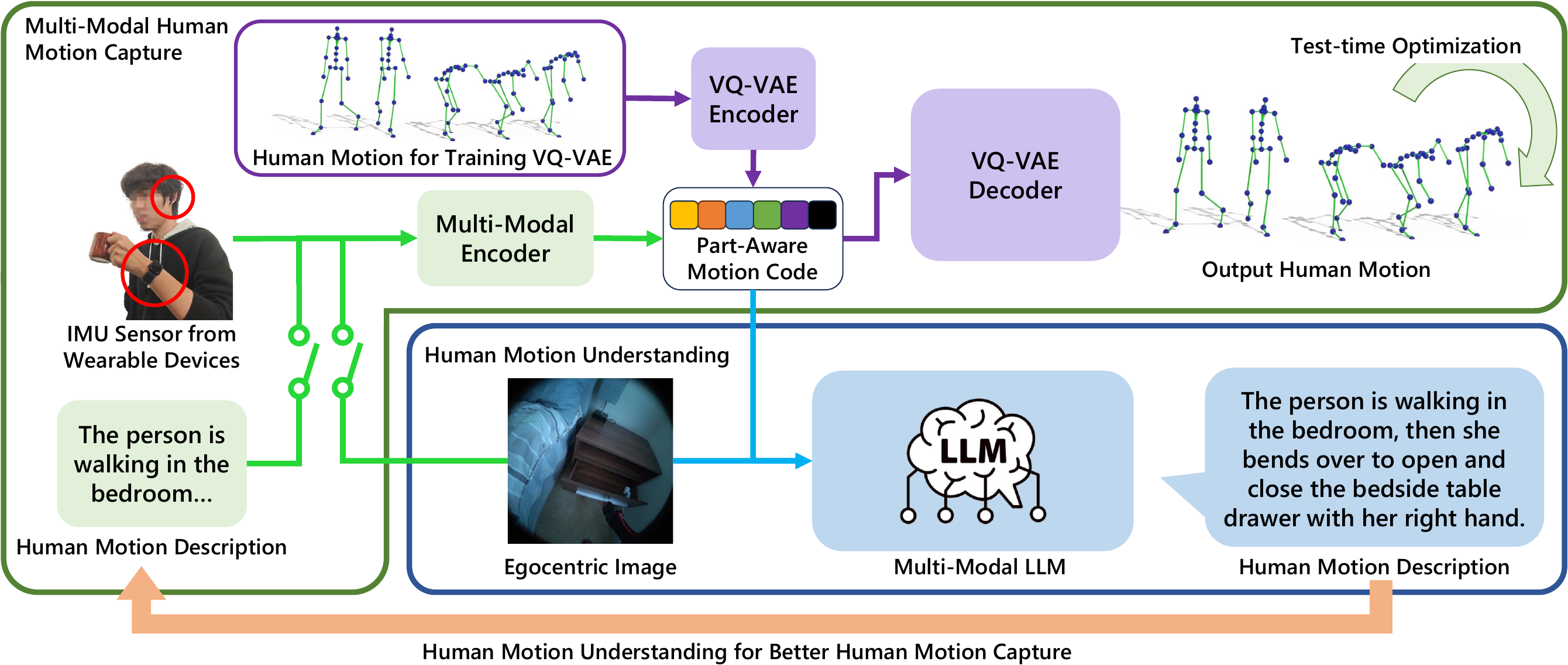}
\caption{
Overview of our Ego4o framework. We first train a VQ-VAE (purple blocks) to learn the part-aware motion codebook (\cref{method:vqvae}). For motion capture (green blocks), the system processes IMU sensor data, egocentric images, and motion descriptions through a multi-modal encoder to generate motion codes in the codebook. These codes are then decoded to predict human motion (\cref{method:mocap}). For motion understanding (blue blocks), the system combines motion codes and egocentric images in a finetuned LLM to generate motion descriptions (\cref{method:motion_understanding}), which can be fed back to enhance motion capture accuracy.
\vspace{-1em}
}
\label{fig:framework}
\end{figure*}
Recent studies in egocentric motion capture have explored multi-modal approaches. EgoLocate~\cite{yi2023egolocate} uses six IMUs and egocentric video for large-area motion capture. EMHI~\cite{fan2024emhi} introduced a dataset combining down-facing stereo cameras, 6DOF trackers, and IMUs, while HMD$^2$~\cite{guzov2024hmd} leverages a conditional diffusion model with egocentric video and head 6DOF pose. However, the input to this method is always fixed.
The work most similar to ours is EgoLM~\cite{hong2024egolm}, which uses head and hand 6D tracking data as input, whereas our method employs 1-3 IMUs. While EgoLM uses a LLM for motion capture, resulting in high computational costs and low accuracy, our approach is faster through a simple encoder-decoder architecture. Moreover, by finetuning on a larger-scale multi-modal LLM, our method enables multi-round conversation and generalization to out-of-distribution images.

\noindent\textbf{Human Motion Generation.}
Human motion generation has been a long-standing challenge in computer vision and graphics. Some works~\cite{petrovich2021action, guo2020action2motion, xu2023actformer} generate human motion from action labels. However, action labels provide only limited representational ability. Recently, numerous works~\cite{zhang2022motiondiffuse, punnakkal2021babel, guo2022generating, zhang2023generating, zhou2025emdm, dabral2023mofusion, tevet2023human, shafir2023human} have focused on generating human motion from text descriptions.
Researchers have also leveraged powerful LLMs to model the joint motion-language distribution~\cite{guo2022tm2t, jiang2023motiongpt, chen2024motionllm, wu2024motionllm}, enabling both human motion generation from text input and text generation from motion. While we use a similar approach to enable human motion understanding with LLMs, our Ego4o framework differs by focusing on accurate human motion capture and supporting multiple egocentric modalities.

\noindent\textbf{Egocentric Motion Understanding.}
Recently, many works have aimed at human motion understanding from the egocentric perspective. Previous works~\cite{damen2018scaling, damen2020epic, sudhakaran2019lsta, lu2019deep, wang2021interactive, li2021trear} usually use egocentric head-mounted front-facing cameras for the human action recognition task. More recently, some researchers~\cite{jia2022egotaskqa, xu2024retrieval, grauman2022ego4d, xue2023egocentric, chen2023egoplan, dessalene2023leap, suglia2024alanavlm} have leveraged Large Language Models (LLMs) for egocentric human motion understanding and have used natural language as output.
In contrast to these methods, our Ego4o approach leverages multi-modal egocentric information, including egocentric images, text descriptions, and IMU signals as input. Our framework can capture human motion and simultaneously generate descriptions about the human activity.

\section{Method}
Our method (Fig.~\ref{fig:framework}) processes a combination of egocentric images $I$, textual motion descriptions $X_a$, and data from one, two, or three IMU sensors, including device acceleration $A$ and rotation $R$. The IMU sensors may be placed at up to five locations: the head, the wrists, or the hips, reflecting typical placements for devices like VR headsets, smartwatches, and cellphones. 
From these multi-modal inputs, we achieve accurate motion capture and can generate textual description of the motion when they are absent.
To achieve this, we first train a part-based motion VQ-VAE (\cref{method:vqvae}) to learn the discrete motion representation for IMUs, then use the multi-modal encoder to project the inputs to the motion representation space(\cref{method:mocap})
The discrete motion codes, while designed for IMU-based motion capture, can also be reused for generating motion descriptions (\cref{method:motion_understanding}).
Finally, we show that motion descriptions generated by our multi-modal LLM can further enhance motion capture performance. (\cref{method:gen_text}). 

\subsection{Learning Part-Aware Motion Representation}~\label{method:vqvae}
In this section, we describe how to learn discrete human motion representation with VQ-VAE~\cite{van2017neural} and further enable the projection of multi-modal inputs to the motion representation space. 
Most previous works~\cite{guo2022tm2t, zhang2023generating, jiang2023motiongpt, chen2024motionllm, zhou2024avatargpt} treat the human body as a holistic entity, encoding the full human body motion into a single VQ-VAE codebook. Though this holistic encoding is effective, it presents limitations for our use case. 

Our method aims to support flexible IMU sensor configurations, ranging from a single head-mounted IMU in smart glasses to various combinations of sensors embedded in smartwatches and smartphones. To achieve this adaptability, we implement the part-aware VQ-VAE architecture introduced in TLControl~\cite{wan2023tlcontrol}, which establishes separate motion codebooks for individual body segments.
These separate motion codebooks enable the direct projection of available IMU signals into their corresponding part-specific motion codebooks, while simultaneously facilitating the generation of latent features for body segments lacking sensor coverage. 
For example, when processing data from a wrist-mounted IMU, the system not only projects this information into the arm-specific motion codebook but also generates leg movements.
This projection mechanism operates analogously to a text-infilling task~\cite{lewis2019bart}, where the system infers motion patterns for unmonitored body segments based on the available sensor data. By employing this part-aware architecture, our system achieves better motion capture accuracy across diverse IMU configurations, offering a versatile solution compared to conventional holistic approaches.


Next, we discuss how to learn the motion representation with part-aware VQ-VAE.
Specifically, all the joints are first divided into six joint groups, including head, left arm, right arm, root, left leg, and right leg. The input ground truth human motion is first encoded to the HumanML3D~\cite{guo2022generating} representation $J\in \mathbb{R}^{T\times M}$, where $T$ is the motion length and $M=263$ corresponds to the motion representation dimensions. Next, $J$ in each time step is split into six groups according to the correlated human body part: $J = [J_\text{Head}, J_\text{LArm}, J_\text{RArm}, J_\text{LLeg}, J_\text{RLeg}, J_\text{Root}]$.
For each body part $i$, we train a separate encoder $\mathcal{E}_i$ to learn an independent codebook $C_i\in \mathbb{R}^{N_\text{code} \times d}$, where $N_\text{code}$ is the size of codebook while $d$ is the dimension of each codebook. The encoder first encodes the human motion into features $Q_i \in \mathbb{R}^{T' \times d}$, where $T' = T/4$. 
Next, the $Q_i$ is quantized with the codebook $C_i$, obtaining the quantized feature $\hat{Q}_i$. The quantized features from all of the body parts are finally concatenated and sent to the VQ-VAE decoder $\mathcal{D}$ to get the reconstructed motion $\hat{J}_\text{recon}$.
The training for the part-aware VQ-VAE is detailed in the suppl. mat.
In the next section, we project multi-modal inputs into this representational space for motion capture and understanding. 

\subsection{Multi-Modal Human Motion Capture}~\label{method:mocap}

In this section, we introduce our multi-modal human motion capture method. The process begins with a transformer-based multi-modal encoder that projects IMU signals, egocentric images, and motion descriptions into the motion representation space learned by the VQ-VAE. These motion features are then processed by the VQ-VAE decoder to reconstruct human motion. Additionally, we offer an optional test-time optimization procedure that can further enhance the accuracy of the motion capture results.

\subsubsection{Multi-Modal Encoder}

Our transformer-based multi-modal encoder processes three input types: motion description $T_m$, egocentric image $I$, and IMU signal sequences. 
The egocentric image $I$ and motion description $T_m$ are encoded into image features $F_I$ and textual features $F_T$ respectively using CLIP~\cite{radford2021learning}. The IMU signal sequence comprises acceleration vectors $A \in \mathbb{R}^{T\times N_{imu} \times 3}$ and rotation matrices $R \in \mathbb{R}^{T\times N_{imu} \times 3 \times 3}$, where $T$ represents the sequence length and $N_{imu} = 5$ is the \textit{maximum} number of IMU locations. In practice, our method is designed to work with arbitrary number of IMUs. 
%
The rotation matrices $R$ are converted to 6D representations~\cite{zhou2019continuity} $R_{6d} \in \mathbb{R}^{T\times N_{imu} \times 6}$.
The IMU acceleration and rotation data are then concatenated and reshaped into an input IMU sequence $F_{imu}$ of length $T'\times N_{imu}$, where $T' = T/4$. 
These IMU sequences $F_{imu}$, along with the image features $F_I$ and the textual motion description features $F_T$, are processed through an embedding layer before being fed into a transformer encoder~\cite{vaswani2023attentionneed}. 
The encoder predicts the logits of the motion code IDs $L_{t, i}$ for the $i^{th}$ IMU at each time step $t$ of the input sequence. Finally, we employ Gumbel Softmax~\cite{jang2016categorical} to get the motion code index $\delta_i \in \{0, 1, 2, ..., N_\text{code}\}$ of the corresponding $i^{th}$ IMU and select the quantized motion feature $\hat{Q}_{t, i}$ from the motion code book. 
The quantized motion features are sent to the VQ-VAE decoder $\mathcal{D}$ following the same way in \cref{method:vqvae} to get the human motion prediction $\hat{J}$. The multi-modal encoder can be trained with the following loss function, which includes the motion code classification loss and human motion reconstruction loss:

\begin{equation} \label{eq:encoder}
    \mathcal{L} = \mathbb{E}_{\hat{L}} \left( -\log P(\hat{L} | A, R, I, T_m) \right) + \lambda \left\| \hat{J} - J \right\|_2
\end{equation}
where $\lambda$ is the weight of reconstruction loss.
To simulate real-world scenarios where certain input modalities may be unavailable, we implement a masking strategy during training. This involves randomly masking egocentric images and textual descriptions. We also simultaneously select random combinations of one to three IMU sensors as active inputs. The remaining IMU sensors are masked to ensure our model learns to operate effectively with varying sensor availability.

\subsubsection{Test-Time Optimization} \label{method:optim}

Limb movement in the motion prediction $\hat{J}$ may not fully align with the corresponding IMU's acceleration and orientation. This can be refined through optional test-time optimization.
The task is to find a motion feature $Q$ in the VQ-VAE latent space such that the reconstructed human motion $J = \mathcal{D}(Q)$, where $\mathcal{D}$ is the \textit{frozen} VQ-VAE decoder, minimizes the energy function:
\begin{equation} \label{eq:optim}
    Q^*sch = \argmin_Q \lambda_a L_a(J, A) + \lambda_r L_r (J, R)
\end{equation}
where $L_a(\cdot)$, $L_r(\cdot)$ are the IMU acceleration term and IMU orientation term, respectively. 
For simplicity, we assume IMUs are positioned near their corresponding body joints—for instance, a smartphone's IMU approximates hip joint motion.
To compute the IMU acceleration term, we first calculate the acceleration of each joint position that corresponds to an IMU sensor placement. For the joint associated with the $i^{th}$ IMU at time step $t$, the acceleration is calculated using: $\hat{\mathbf{a}}^i_t = (J^i_{t+2} - 2J^i_{t+1} + J^i_t)/(\Delta t^2)$, where $\Delta t = 1$ in our experiment. 
The overall IMU acceleration term is calculated as: $L_a(J, A) =\sum_i\sum_t\| \hat{\mathbf{a}}^i_t - \mathbf{a}^i_t \|_2$, where $t = 0, 1, 2, ...$ represents time steps in the motion sequence, $i$ indexes the available IMU sensors.

To compute the IMU orientation term, we first calculate the orientation of each limb in predicted motion $\hat{J}$ that corresponds to an IMU sensor placement. For the limb associated with the $i$th IMU at time step $t$, the orientation vector is: $\hat{\mathbf{r}}^i_t = (J^i_{t, \text{child}} - J^i_{t, \text{parent}}) / \|J^i_{t, \text{child}} - J^i_{t, \text{parent}}\|_2$
where $J^i_{t, \text{child}}$ and $J^i_{t, \text{parent}}$ represent the child and parent joint positions of the limb segment associated with the $i$th IMU.

Next, we calculate the orientation of each available IMU sensor with $\mathbf{r}^i_t = M^i \cdot R^i_t \cdot [0, 1, 0]^T$
where $R^i_t$ represents the rotation matrix of the $i$th IMU at time step $t$, $[0, 1, 0]^T$ denotes the initial orientation vector, and $M^i$ represents the calibration rotation matrix between the IMU sensor and its corresponding limb segment, determined through prior calibration. The IMU orientation term is then computed as:
$L_r(J, R) = \sum_i\sum_t\|\hat{\mathbf{r}}^i_t - \mathbf{r}^i_t\|_2$, where $i$ indexes the available IMU sensors.

\subsection{Egocentric Human Motion Understanding}~\label{method:motion_understanding}

In this section, we present our approach to human motion understanding through multi-modal LLM fine-tuning. While existing pre-trained multi-modal LLMs excel at modeling language and image distributions, they lack the capability to process data related to human motion. 
To natively enable such understanding of human motion, we extend LLaVA (Vicuna-7B)~\cite{liu2024visual} by incorporating a new motion modality and fine-tuning it using the multi-modal egocentric dataset Nymeria~\cite{ma2024nymeria}. 

\subsubsection{Architecture}

\begin{figure}
\centering
\includegraphics[width=1\linewidth]{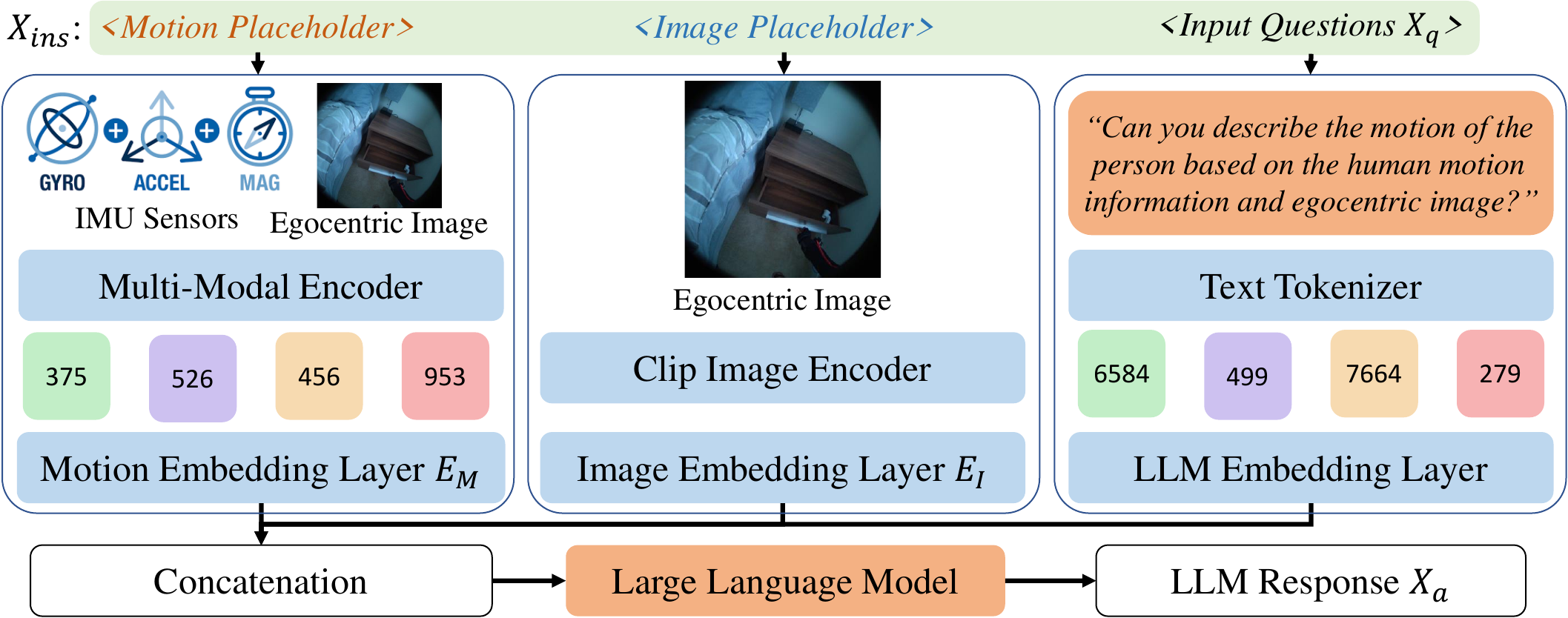}
\caption{
Egocentric Human Motion Understanding. Each modality is encoded separately and then concatenated in the order specified by the input instruction $X_\text{ins}$ before being fed into the LLM.
\vspace{-1em}
}
\label{fig:llm}
\end{figure}

The architecture of our egocentric LLM is shown in \cref{fig:llm}.
Given an input image $I$ and IMU sensor data $(A, R)$, we first employ the multi-modal encoder described in \cref{method:mocap} to generate human motion codes, which serve as discrete motion tokens. 
These motion codes are then processed through a linear embedding layer $\mathbf{E}_M$ to produce human motion features $H_M$, aligned with the language model's word embedding dimensionality. 
%
%
For image processing, we utilize a pretrained CLIP~\cite{radford2021learning} image encoder $E_I$ to map image features $F_I$ into the word embedding space, resulting in $H_I$. 
Finally, the image features $H_I$, motion features $H_M$, and text encodings $H_T$ are concatenated and fed into the LLM to generate the response.

\subsubsection{Training}
For each IMU signal sequence $(A, R)$ and the corresponding egocentric image $I$, we generate conversation pairs $(X_q, X_a)$. The questions $X_q$ (see an example in \cref{fig:llm}) are prompts requesting human motion descriptions, randomly sampled from a pre-defined list. The answers $X_a$ are drawn from the fine-grained motion descriptions in the Nymeria~\cite{ma2024nymeria} dataset. The input instruction set is constructed as $X_\text{ins} = \operatorname{RandomSelect} \{ [I, X_q], [A, R, X_q], [A, R, I, X_q] \}$.

The LLM can be fine-tuned on prediction tokens using an auto-regressive training objective. The probability of generating the answer $X_a$ is computed as:
\begin{equation}
p(X_a | X_\text{ins}) = \prod_{i=1}^{L} p_{\theta}(x_i | X_v, X_{\text{ins}, <i}, X_{a, <i})
\end{equation}
where $L$ represents the token sequence length, $\theta$ denotes the trainable parameters, and $X_{\text{ins},<i}$ and $X_{a,<i}$ are the instruction and answer tokens preceding the current prediction token $x_i$. The trainable parameters are optimized using the negative log-likelihood loss.

The training process consists of two stages: in the first stage, we conduct motion pre-training to achieve motion feature alignment. In the second stage, we use multi-modal fine-tuning to enable egocentric motion understanding. 

\textbf{Motion Pre-Training.}
For the pre-training phase, we restrict input instructions to those containing only IMU signals $[A, R, X_q]$. To ensure proper alignment between motion features and the pre-trained LLM's word embeddings, we exclusively train the motion embedding layer $E_M$ while keeping all other architectural components frozen.

\textbf{Multi-Modal Finetuning.}
In this phase, we maintain frozen weights for both the CLIP encoder and multi-modal encoder, while continuing to update three components: the image embedding layer $E_I$, motion embedding layer $E_M$, and LLM parameters using LoRA~\cite{hu2021lora} finetuning.

\subsection{Ego4o-LLM Descriptions for Better MoCap}\label{method:gen_text}

While our motion capture module functions effectively without verbal descriptions, incorporating high-quality motion descriptions can significantly enhance its accuracy, especially in disambiguating the challenging cases caused by self-occlusion. 
However, obtaining such descriptions in real-world scenarios presents a challenge, as users are typically reluctant to narrate their actions in real-time. 
To address this limitation, we leverage our system's ability to generate accurate motion descriptions through the Ego4o LLM. 
Though these generated descriptions may not perfectly match human-provided reference descriptions, they prove valuable inductive bias for enhancing motion capture performance. 
To further bridge the gap between generated and ground truth descriptions, we finetune our multi-modal encoder using generated descriptions for only 300 iterations, with results detailed in our ablation study (\cref{exp:ablation}).
\section{Experiments}

\subsection{Datasets and Evaluation Metrics}

\noindent\textbf{Datasets}
In our experiments, we evaluate our method on two datasets: the DIP-IMU dataset~\cite{huang2018deep} for assessing human motion capture accuracy from IMU devices, and the Nymeria dataset~\cite{ma2024nymeria} for evaluating both motion capture accuracy and motion description generation quality.
For the results on DIP-IMU dataset, we first train the VQ-VAE on the AMASS dataset~\cite{mahmood2019amass}, then train it on synthetic IMU-based motion capture data generated from AMASS following IMUPoser~\cite{mollyn2023imuposer}. The network is subsequently fine-tuned on the DIP-IMU training split before evaluation.

The Nymeria dataset contains approximately 170k human motion sequences, each 5 seconds in duration. We split the sequences into training ($\sim$ 119k sequences) and test ($\sim$ 51k sequences) sets based on different scenes and motion capture identities. For the evaluation on the Nymeria dataset, we train the VQ-VAE and multi-modal encoder and fine-tune the Ego4o LLM on the Nymeria training dataset. More implementation and training details are in suppl. mat.

\begin{figure*}
\centering
\includegraphics[width=0.95\linewidth]{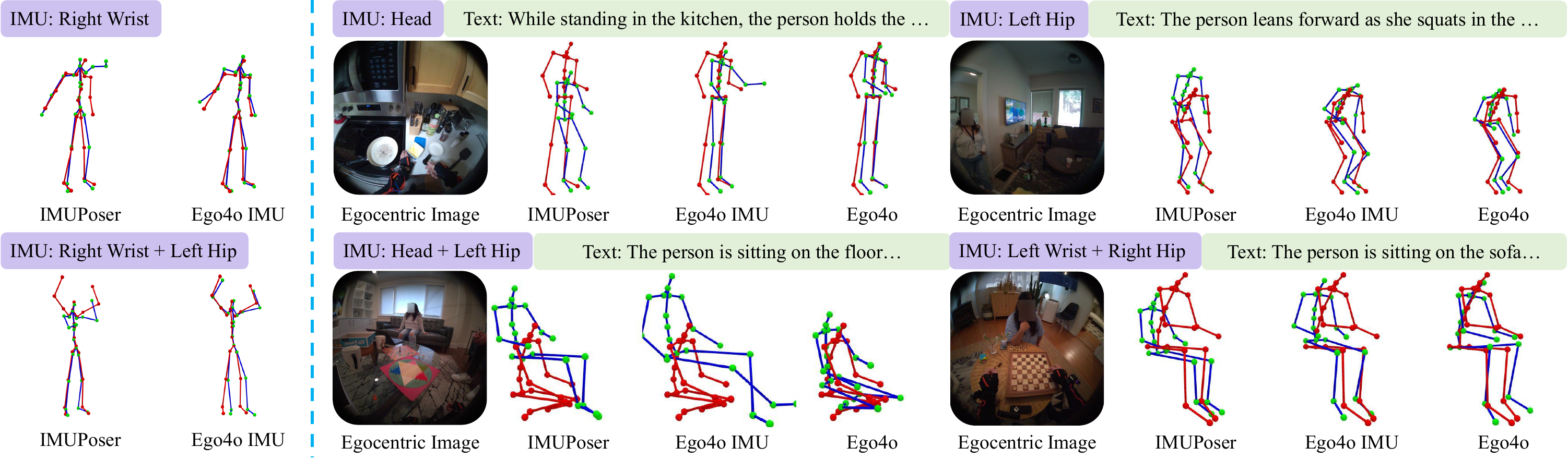}
\caption{
Comparison of human motion capture results between Ego4o, Ego4o-IMU and IMUPoser~\cite{mollyn2023imuposer} on the DIP-IMU~\cite{huang2018deep} (left) and Nymeria dataset~\cite{ma2024nymeria} (right). The red skeleton is the ground truth, while the green skeleton is the predicted pose. Our predictions are more accurate than the baselines when only using IMU input, and using egocentric images and motion descriptions improves the performance.
\vspace{-1em}
}
\label{fig:result_mocap}
\end{figure*}

\noindent\textbf{Evaluation Metrics}
For evaluating human motion capture accuracy, we calculate joint position errors using MPJPE and PA-MPJPE (with Procrustes alignment). We also evaluate joint jitter error to assess the smoothness of predicted motion. For motion understanding, which generates natural language outputs, we employ NLP metrics including BERT score~\cite{zhang2019bertscore}, BLEU~\cite{papineni2002bleu}, and ROUGE-L~\cite{lin2004rouge}. Details of the evaluation metrics are provided in the suppl. mat.

\subsection{Comparisons on IMU-Based Human Mocap}
In this section, we present our egocentric human motion capture results. Since no publicly available method supports as many modalities, we compare Ego4o to the most relevant IMU-based human motion capture IMUPoser ~\cite{mollyn2023imuposer} on the DIP-IMU~\cite{huang2018deep} and Nymeria~\cite{ma2024nymeria} datasets. For a fair comparison, we disable the egocentric image and motion description inputs and use only 1-3 IMUs, naming this setup Ego4o-IMU. We also evaluate the full multimodal Ego4o. 

\begin{figure}
\centering
\includegraphics[width=0.92\linewidth]{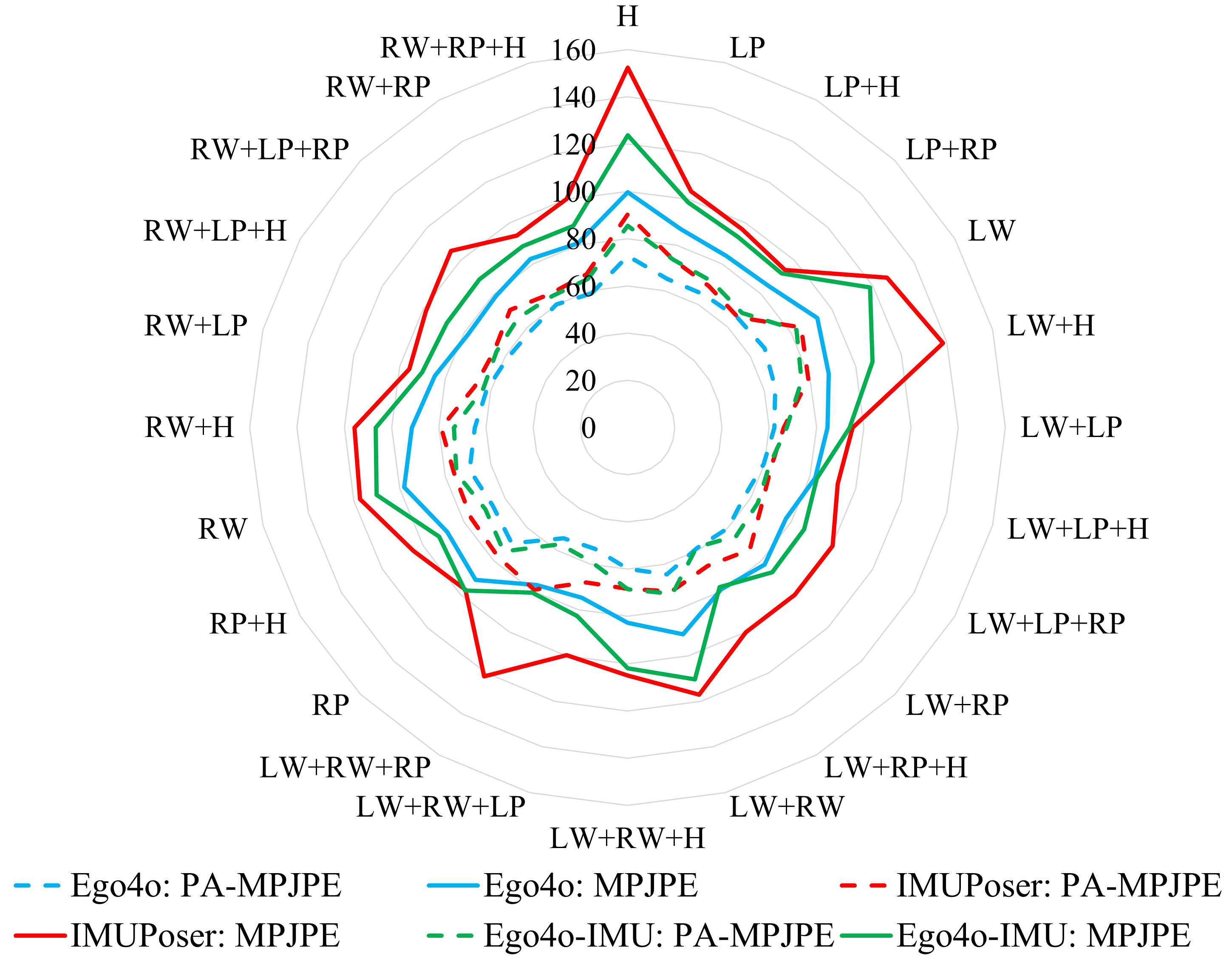}
\caption{
Quantitative results of human motion capture on Nymeria dataset. The result compares our method with IMUPoser under different IMU setups. H, LP, RP, LW, and RW indicate the IMU located on different body parts. H: head, LP: left hip, RP: right hip, LW: left wrist, RW: right wrist. 
\vspace{-1em}
}
\label{fig:result_mocap_radar}
\end{figure}

The results in \cref{fig:result_mocap_radar} show the performance of Ego4o, Ego4o-IMU, and IMUPoser under different IMU setups on the Nymeria dataset, following the same evaluation protocol as IMUPoser. The results in \cref{table:mocap_results} present the average performance across these various IMU configurations. These results demonstrate that Ego4o outperforms IMUPoser when using only IMU inputs. Furthermore, incorporating egocentric images and motion descriptions further enhances Ego4o's performance. Unfortunately, we were unable to compare against some related works~\cite{xu2024mobileposer, van2024diffusionposer} due to the lack of available code.
For a qualitative comparison, we visualize the body poses estimated by Ego4o and IMUPoser on the DIP-IMU and Nymeria datasets in \cref{fig:result_mocap}. Results show that Ego4o method can accurately predict human pose from not only IMU sensor inputs but also the multimodal inputs of egocentric images and motion descriptions.

\begin{table}
\begin{center}
\small
\begin{tabularx}{0.47\textwidth} { 
   >{\raggedright\arraybackslash}X 
   >{\centering\arraybackslash\hsize=0.6\hsize}X 
   >{\centering\arraybackslash\hsize=0.6\hsize}X
   >{\centering\arraybackslash\hsize=0.6\hsize}X}
\hlineB{2.5}
Method & MPJPE (mm) & PA-MPJPE (mm) & Jitter ($km/s^3$) \\
\hline
\multicolumn{2}{l}{\textbf{DIP-IMU Dataset}} \\
\hline
DIP (6 IMU) &  73 & -- & 3.01  \\
TransPose (6 IMU) &  59 & -- & 0.14   \\
\hline
IMUPoser &  97 & -- & 0.19   \\
Ego4o-IMU & \textbf{84.06} & \textbf{63.95} & \textbf{0.076}  \\
\hline
\multicolumn{2}{l}{\textbf{Nymeria Dataset}} \\
\hline
IMUPoser &  105.7 & 72.94 & 0.054   \\
Ego4o-IMU &  \underline{95.86} & \underline{69.03} & \underline{0.049}  \\
Ego4o &  \textbf{84.82} & \textbf{62.33} & \textbf{0.048}  \\
\hlineB{2.5}
\end{tabularx}
\end{center}
\vspace{-1em}
\caption{
Quantitative results for human motion capture: On DIP-IMU, 1-3 IMUs were used. For Nymeria, IMUPoser and Ego4o-IMU used 1-3 IMUs, while Ego4o utilized 1-3 IMUs, a single egocentric image, and ground truth motion descriptions.
\vspace{-1.5em}
}
\label{table:mocap_results}
\end{table}

\begin{figure*}
\centering
\includegraphics[width=0.96\linewidth]{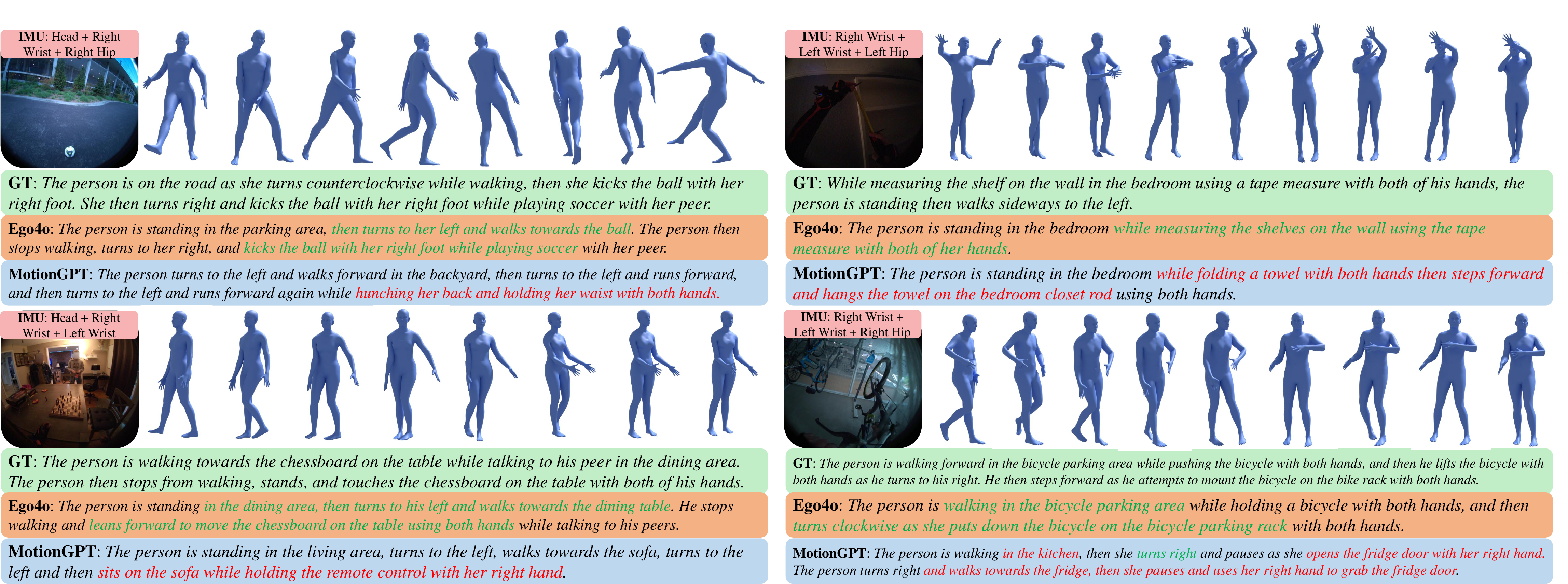}
\caption{
Comparison of motion description generation between Ego4o and MotionGPT~\cite{jiang2023motiongpt}. The egocentric image and ground truth human motion (for reference) are shown. Highlight predictions are marked in green, while incorrect predictions are in red. Ego4o's descriptions are more accurate, demonstrating the benefits of its joint modeling of multimodal inputs.
\vspace{-1em}
}
\label{fig:result_text}
\end{figure*}

\subsection{Comparisons on Motion Understanding}

This section highlights Ego4o's motion description generation capabilities. In this experiment, we do not use the motion description as input and instead rely solely on random 1-3 IMU sensors and egocentric images. We compare Ego4o's performance against previous motion description generation methods, TM2T~\cite{guo2022tm2t} and MotionGPT~\cite{jiang2023motiongpt}, where we first predict the human motion using Ego4o and then use those predictions as input to the other networks.
The comparison results are shown in \cref{table:text_results}, where Ego4o outperforms the previous methods across most metrics, particularly in terms of BERTScore~\cite{zhang2019bertscore} and RougeL~\cite{lin2004rouge}. Ego4o did not surpass the prior methods in Bleu@4, as the Bleu score~\cite{papineni2002bleu} focuses solely on n-gram overlap and cannot fully capture the quality of semantic understanding~\cite{zhang2019bertscore}.

\begin{table}
\begin{center}
\small
\begin{tabularx}{0.47\textwidth} { 
   >{\raggedright\arraybackslash}X 
   >{\centering\arraybackslash\hsize=.8\hsize}X 
   >{\centering\arraybackslash\hsize=.8\hsize}X
   >{\centering\arraybackslash\hsize=.8\hsize}X
   >{\centering\arraybackslash\hsize=.8\hsize}X
   >{\centering\arraybackslash\hsize=.8\hsize}X}
\hlineB{2.5}
Method & Bert(idf) & Bleu@1 & Bleu@4  & RougeL \\
\hline
TM2T & 11.08  & 40.11 &  8.99   & 30.70  \\
MotionGPT & 14.09  & 42.22  & \textbf{10.31}    & 32.33 \\
Ego4o & \textbf{30.13} & \textbf{53.83} & 7.46  & \textbf{38.95} \\
\hlineB{2.5}
\end{tabularx}
\end{center}
\vspace{-1em}
\caption{Quantitative results of motion description generation.
\vspace{-1em}
}
\label{table:text_results}
\end{table}

\subsection{Ablation Study}\label{exp:ablation}

\subsubsection{Ablation Study on Human Motion Capture} 

\noindent\textbf{Test-time optimization.} To assess the performance impact of our test-time optimization (\cref{method:optim}), we include results by not using the test-time optimization as ``w/o optim'' in \cref{table:ablation_study_mocap}. The MPJPE scores are higher, demonstrating the effectiveness of this module. We notice that the jitter is smaller when not using the test-time optimization. This is caused by the optimization with noisy IMU signals.

\noindent\textbf{Multi-modal input.} We evaluate the impact of different input modalities on motion capture performance. When evaluating the ``only gt text'' case in \cref{table:ablation_study_mocap}, which only uses the ground truth motion description and IMUs as input, and the ``only image'' case in \cref{table:ablation_study_mocap}, which only uses the egocentric image and IMUs, the results show a significant drop in performance compared to the full multi-modal setup. This highlights the complementary value that both the egocentric image and the ground truth motion description bring to enhancing the model's motion capture capabilities. 

\begin{table}
\begin{center}
\small
\begin{tabularx}{0.47\textwidth} { 
   >{\raggedright\arraybackslash}X
   >{\centering\arraybackslash\hsize=.62\hsize}X 
   >{\centering\arraybackslash\hsize=.62\hsize}X
   >{\centering\arraybackslash\hsize=.62\hsize}X}
\hlineB{2.5}
Method  & MPJPE (mm) & PA-MPJPE (mm) & Jitter ($km/s^3$) \\
\hline
Ego4o-IMU & 95.86 & 69.03 & 0.049  \\
\hline
w/o optim & 85.93 & 64.02 & 0.039 \\
only gt text & 86.22 & 63.14 & 0.048  \\
only image & 90.81 &  66.04 & 0.049  \\
w/ gen text & 88.65 & 64.79 & 0.048 \\
image\&gen text & 87.00 & 63.67 & 0.049 \\
\hline
Ego4o & 84.82 & 62.33 & 0.048  \\
\hlineB{2.5}
\end{tabularx}
\end{center}
\vspace{-1em}
\caption{Ablation study of the human mocap on Nymeria dataset. 
\vspace{-1em}
}
\label{table:ablation_study_mocap}
\end{table}

\noindent\textbf{Generated text for better motion capture.} 
In \cref{method:gen_text}, we claim that if the ground truth motion description is unavailable, utilizing the generated text as input to the human motion capture module could enhance performance. To demonstrate this, in this experiment, we evaluate the motion capture performance under two settings: First, with the generated motion description and IMUs as input, the results are shown as ``w/ gen text''. This performance is better than the ``Ego4D-IMU'' result, which only uses IMU data, and slightly worse than the ``only gt text'' case, which takes the ground truth text and IMU as input.
Second, with the generated description, IMUs, and an egocentric image as input, the results are shown as ``image\&gen text''. This performs better than the ``only image'' case, which uses the egocentric image and IMUs, and slightly worse than our full method.

The results demonstrate that using the generated descriptions, even if they do not perfectly match the ground truth, still leads to notable improvements in motion capture performance compared to the no-text baseline. This highlights the value of utilizing generated text to enhance the system's capabilities when ground truth descriptions are unavailable.


\begin{table}
\begin{center}
\small
\begin{tabularx}{0.47\textwidth} { 
   >{\raggedright\arraybackslash}X 
   >{\centering\arraybackslash\hsize=.8\hsize}X 
   >{\centering\arraybackslash\hsize=.7\hsize}X
   >{\centering\arraybackslash\hsize=.7\hsize}X
   >{\centering\arraybackslash\hsize=.7\hsize}X
   >{\centering\arraybackslash\hsize=.7\hsize}X}
\hlineB{2.5}
Method & Bert(idf) & Bleu@1 & Bleu@4 & RougeL \\
\hline
w/o image & 22.44 & 48.63 & 5.81  & 36.48  \\
w/o motion & 25.55 & 50.90 & 6.32  & 35.71 \\
gt motion & 31.38 & 54.78 & 9.44 & 39.86 \\
\hline
Ego4o & 30.13 & 53.83 & 7.46  & 38.95 \\
\hlineB{2.5}
\end{tabularx}
\end{center}
\vspace{-1em}
\caption{Ablation of motion understanding on Nymeria dataset. 
\vspace{-1em}
}
\label{table:text_ablation}
\end{table}

\subsubsection{Ablation Study on Human Motion Understanding}
\noindent\textbf{Multi-modal input.} In this experiment, we evaluate the performance of motion description generation without the egocentric image or without a human motion token as input. The results without image input, labeled as ``w/o image'' in \cref{table:text_ablation} show a noticeable decline across all metrics. Without image context, the language model loses key contextual information, leading to reduced accuracy in motion description.
The results without the human motion token input are shown as ``w/o motion'' in \cref{table:text_ablation}. The absence of human motion information causes a performance drop, as the egocentric image cannot see the human body, making it difficult to generate an accurate motion description.

\noindent\textbf{Ground truth human motion.} From ``gt motion'' row in \cref{table:text_ablation}, the accuracy of motion description generation is enhanced by using the ground truth motion codes (encoded by VQ-VAE encoder), compared to our Ego4o method that uses encoded motion tokens from IMUs and egocentric images. This suggests motion information is important in the model's understanding and generation capabilities.

%
\section{Discussion}
\noindent \textbf{Limitations.}
Despite outperforming the state-of-the-art in various evaluation scenarios, our method has a few practical limitations.
First, it requires motion sequences as input, which introduces latency in online applications. 
Second, the system's capacity for multi-round conversational interaction remains limited. 
To address this, future work can use better instructional fine-tuning of the large language model to generate multi-round conversational datasets.

\noindent \textbf{Conclusion.}
In this paper, we introduced Ego4o, a framework for egocentric human motion capture and understanding that combines multi-modal inputs from wearable devices. 
Our versatile design allows us to operate not only with a variable number of IMUs, but also can optionally incorporate text and images.
By integrating kinematic data and semantic information, Ego4o achieves high accuracy in motion capture while providing detailed motion descriptions.
We also showed text descriptions generated by a motion-aware LLM can in turn be used to perform better text-assisted motion capture. 
Our experiments demonstrate significant improvements in both tracking accuracy and description quality compared to existing methods.
%
We envision future extensions of this work toward a human foundational model that adapts to various sensing modalities, incorporates common-sense reasoning about human attributes, and interacts naturally with users via text/audio.

\noindent\textbf{Acknowledgment} The work was supported by Saarbrücken Research Center for Visual Computing, Interaction and AI.

{
    \small
    \bibliographystyle{ieeenat_fullname}
    \bibliography{main}
}
\clearpage
\setcounter{page}{1}

\maketitlesupplementary

\begin{table*}[t]
\begin{center}
\small
\begin{tabular}{lcccccccccc}
\toprule
Setups & \multicolumn{2}{c}{IMUPoser~\cite{mollyn2023imuposer}} & \multicolumn{2}{c}{Ego4o-IMU} & \multicolumn{2}{c}{Ego4o} \\
\cmidrule(r){2-3} \cmidrule(r){4-5} \cmidrule(r){6-7}
 & MPJPE & PA-MPJPE & MPJPE & PA-MPJPE & MPJPE & PA-MPJPE \\
\midrule
        H & 90.34 & 152.5 & 85.45 & 123.8 & 72.60 & 99.68 \\
        LP & 73.65 & 103.6 & 73.73 & 98.78 & 65.06 & 86.94 \\
        LP+H & 69.05 & 97.02 & 71.44 & 93.26 & 65.00 & 83.84 \\
        LP+RP & 66.08 & 94.27 & 68.62 & 92.27 & 65.90 & 84.56 \\
        LW & 84.80 & 126.9 & 82.59 & 118.5 & 67.05 & 92.75 \\
        LW+H & 79.27 & 138.3 & 76.20 & 107.4 & 64.80 & 88.19 \\
        LW+LP & 66.05 & 95.31 & 67.41 & 94.05 & 62.18 & 84.73 \\
        LW+LP+H & 63.12 & 92.06 & 62.16 & 83.09 & 59.66 & 82.25 \\
        LW+LP+RP & 65.63 & 100.2 & 63.73 & 86.36 & 57.48 & 77.36 \\
        LW+RP & 73.21 & 100.1 & 65.07 & 86.71 & 60.12 & 81.98 \\
        LW+RP+H & 67.66 & 100.1 & 58.49 & 77.98 & 58.95 & 79.17 \\
        LW+RW & 71.74 & 117.1 & 73.02 & 110.4 & 64.38 & 90.74 \\
        LW+RW+H & 68.32 & 105.0 & 68.42 & 101.9 & 59.78 & 82.73 \\
        LW+RW+LP & 67.93 & 99.80 & 59.01 & 82.62 & 53.25 & 74.61 \\
        LW+RW+RP & 79.26 & 121.6 & 56.96 & 80.71 & 54.10 & 77.05 \\
        RP & 77.55 & 97.16 & 74.18 & 97.54 & 69.37 & 91.18 \\
        RP+H & 76.66 & 104.6 & 69.48 & 92.30 & 66.08 & 88.33 \\
        RW & 76.01 & 117.3 & 75.01 & 110.2 & 69.21 & 97.98 \\
        RW+H & 79.06 & 115.7 & 73.65 & 106.8 & 64.69 & 91.40 \\
        RW+LP & 67.06 & 95.73 & 63.34 & 90.16 & 61.73 & 84.58 \\
        RW+LP+H & 65.12 & 98.58 & 64.24 & 88.39 & 58.81 & 78.35 \\
        RW+LP+RP & 70.46 & 105.9 & 65.51 & 88.67 & 57.89 & 78.87 \\
        RW+RP & 65.37 & 93.78 & 64.18 & 88.72 & 60.28 & 82.38 \\
        RW+RP+H & 67.24 & 100.1 & 65.19 & 88.46 & 58.89 & 80.82 \\
\bottomrule
\end{tabular}
\caption{Result of the IMU-based human motion capture on the Nymeria Dataset under different IMU setups. H, LP, RP, LW, and RW indicate the IMU located on different body parts. H: head, LP: left hip, RP: right hip, LW: left wrist, RW: right wrist. The results are shown in millimeters.}
\label{table:result_each_imu}
\end{center}
\end{table*}

\begin{table*}[t]
\begin{center}
\small
\begin{tabular}{lcccccccccc}
\toprule
Setups & \multicolumn{2}{c}{w/o optim} & \multicolumn{2}{c}{only gt text} & \multicolumn{2}{c}{only image} & \multicolumn{2}{c}{w/ gen text} & \multicolumn{2}{c}{image \& gen text} \\
\cmidrule(r){2-3} \cmidrule(r){4-5} \cmidrule(r){6-7} \cmidrule(r){8-9} \cmidrule(r){10-11}
 & MPJPE & PA-MPJPE & MPJPE & PA-MPJPE & MPJPE & PA-MPJPE & MPJPE & PA-MPJPE & MPJPE & PA-MPJPE \\
\midrule
H & 99.69 & 72.60 & 95.09 & 70.52 & 106.62 & 78.69 & 104.92 & 75.24 & 96.53 & 71.56 \\
LP & 86.02 & 65.11 & 94.56 & 70.67 & 92.82 & 73.83 & 91.60 & 69.34 & 89.30 & 66.33 \\
LP+H & 85.92 & 65.10 & 84.66 & 67.52 & 87.47 & 67.50 & 85.93 & 65.49 & 86.41 & 66.64 \\
LP+RP & 85.59 & 65.95 & 87.82 & 66.25 & 88.50 & 66.87 & 87.41 & 66.43 & 92.53 & 67.57 \\
LW & 93.76 & 67.06 & 98.22 & 69.93 & 105.46 & 72.68 & 99.22 & 69.62 & 102.66 & 72.69 \\
LW+H & 89.24 & 66.81 & 84.50 & 61.30 & 96.04 & 67.86 & 96.04 & 69.23 & 90.30 & 64.79 \\
LW+LP & 85.73 & 64.21 & 86.83 & 65.24 & 88.02 & 63.25 & 83.79 & 61.25 & 82.68 & 61.52 \\
LW+LP+H & 83.25 & 61.76 & 78.71 & 56.10 & 83.20 & 61.32 & 81.27 & 61.88 & 84.40 & 62.98 \\
LW+LP+RP & 78.45 & 59.53 & 76.78 & 57.25 & 80.98 & 61.06 & 81.55 & 62.33 & 79.40 & 58.54 \\
LW+RP & 83.03 & 62.19 & 82.08 & 61.53 & 88.56 & 65.89 & 89.21 & 69.14 & 84.13 & 63.20 \\
LW+RP+H & 80.18 & 61.04 & 79.53 & 58.29 & 86.23 & 63.73 & 78.52 & 58.45 & 80.57 & 60.19 \\
LW+RW & 91.78 & 66.39 & 86.97 & 60.79 & 100.60 & 70.83 & 97.83 & 66.98 & 95.32 & 66.37 \\
LW+RW+H & 83.75 & 61.86 & 85.38 & 59.33 & 88.71 & 62.52 & 85.05 & 59.64 & 87.50 & 62.34 \\
LW+RW+LP & 75.67 & 55.33 & 86.49 & 61.75 & 83.67 & 59.28 & 81.56 & 59.34 & 79.21 & 57.23 \\
LW+RW+RP & 78.08 & 56.14 & 82.35 & 58.28 & 83.31 & 56.80 & 82.68 & 56.54 & 79.83 & 57.62 \\
RP & 92.25 & 71.45 & 95.32 & 73.19 & 91.01 & 68.26 & 87.94 & 66.35 & 88.91 & 69.01 \\
RP+H & 89.33 & 68.12 & 84.98 & 64.51 & 84.52 & 64.21 & 90.82 & 69.30 & 83.56 & 65.76 \\
RW & 99.01 & 71.26 & 97.44 & 69.03 & 112.56 & 74.99 & 103.29 & 72.26 & 104.32 & 71.15 \\
RW+H & 92.42 & 66.75 & 95.17 & 66.01 & 99.42 & 69.64 & 90.04 & 64.11 & 94.27 & 66.67 \\
RW+LP & 85.64 & 63.77 & 80.35 & 59.87 & 91.61 & 65.31 & 90.86 & 66.36 & 81.71 & 60.36 \\
RW+LP+H & 79.42 & 60.85 & 79.57 & 57.55 & 85.03 & 64.21 & 74.61 & 56.42 & 83.25 & 59.26 \\
RW+LP+RP & 79.90 & 59.93 & 77.60 & 57.26 & 83.17 & 59.37 & 84.70 & 60.58 & 78.54 & 56.40 \\
RW+RP & 83.42 & 62.29 & 86.20 & 62.59 & 86.47 & 63.91 & 88.75 & 64.07 & 82.57 & 61.11 \\
RW+RP+H & 81.84 & 60.93 & 82.52 & 60.57 & 86.44 & 62.98 & 84.56 & 60.68 & 80.13 & 58.59 \\
\bottomrule
\end{tabular}
\caption{Ablation study of the IMU-based human motion capture on the Nymeria Dataset under different IMU setups. H, LP, RP, LW, and RW indicate the IMU located on different body parts. H: head, LP: left hip, RP: right hip, LW: left wrist, RW: right wrist. The results are shown in millimeters.}
\label{table:ablation_study_each_imu}
\end{center}
\end{table*}

\section{Implementation Details}

In this section, we describe the implementation details of our method.

\subsection{Part-Aware VQ-VAE}

\subsubsection{Network Structure}

We first introduce the network structure of the encoder $\mathcal{E}$ for each human body part.
For each joint encoder, we utilize a codebook containing 4096 code vectors, each with a dimension of 64. The input human motion for a specific body part is $J_i \in \mathbb{R}^{T \times D}$, where $T$ represents the motion length and $D$ denotes the dimension of HumanML3D~\cite{guo2022generating} representation. The input first traverses through a 1D convolutional layer (kernel size=3, stride=1, padding=1), followed by a ReLU activation function, producing a feature with 512 channels.

The motion feature then passes through two down-sampling blocks. Each down-sampling block comprises a 1D convolutional layer (kernel size=4, stride=2, padding=1) and three Resnet blocks. A Resnet block consists of a sequential structure: a convolutional layer, followed by a ReLU activation function, and another convolutional layer. The output from this sequence is combined with the input through addition to form the Resnet block's output.

A final 1D convolutional layer (kernel size=3, stride=1, padding=1) is applied to generate the feature $Q_i \in \mathbb{R}^{T' \times d}$, where $d$ equals 64 (matching the code dimension) and $T' = T/4$. Before quantization, the encoded feature undergoes normalization. The full-body latent code $\hat{Q}_i$ is constructed by combining the quantized codes from all six joint encoders.

The decoder mirrors the encoder's architecture, with one key modification: convolutional layers having stride=2 are replaced with upsampling layers using nearest neighbor interpolation. This process finally yields the reconstructed human motion $\hat{J}_\text{recon}$.

\subsubsection{Training Details}

For training the part-aware VQ-VAE, we use standard loss terms including quantization, commitment, and reconstruction losses. 

\begin{equation}
\begin{aligned}
    \mathcal{L} = \sum_{i} & \left( \beta \| \text{sg}[\hat{Q}_i] - Q_i \|_2 \right. \\
& + \left. \| \hat{Q}_i - \text{sg}(Q_i) \|_2 \right)  + \| J - \hat{J}_\text{recon} \|_2
\end{aligned}
\end{equation}
where $\beta$ is a balancing term, $\text{sg}[\cdot]$ denotes the stop-gradient operator.
During training, we employ the Adam optimizer~\cite{kingma2014adam} with a batch size of 128 and a learning rate of $1\times10^{-4}$.

\subsection{Multi-Modal Encoder}

\subsubsection{Network Structure}

In implementing the masked trajectory transformer, we utilize a pre-trained CLIP-ViT-B/32~\cite{radford2021learning} model to extract features from the egocentric image and motion description. The IMU signals $(A, R)$ are grouped into single feature tokens, with each token spanning 4 time steps. We do this grouping since it aligns with our downsampling rate of 4 in the part-based VQ-VAE framework. Each token transforms into a 512-dimensional feature through a linear projection layer.

The image features, motion description features, and IMU tokens are then concatenated and processed through a 4-layer transformer encoder to obtain the latent space representation. In each transformer encoder layer, the attention head number is 4, the dimension of the feed-forward network is 2048, and the dropout rate is 0.1. Subsequently, a 3-layer transformer encoder transforms this latent space into a sequence of logits.  In each transformer encoder layer, the attention head number is 4, the dimension of the feed-forward layer is 1024 and the dropout rate is 0.1. We employ GumbelSoftmax~\cite{jang2016categorical} to convert these logits into motion code indices $\delta_i$ for each possible IMU location $i$. The final motion features $\hat{Q}_i$ are obtained by selecting from the corresponding VQ-VAE codebook $C_i$.

\subsubsection{Training Details}

For training the multi-modal encoder, we optimize the encoder network while keeping the VQ-VAE decoder and CLIP~\cite{radford2021learning} model frozen. The network is trained for 25 epochs using the Adam optimizer~\cite{kingma2014adam} with a learning rate of $1 \times 10^{-4}$ and a batch size of 128. During training, the weighting parameter $\lambda$ in \cref{eq:encoder} is set to 0.001.

\subsubsection{Optimization Details}
In the energy function \cref{eq:optim} in \cref{method:optim}, we set the weights $\lambda_a = 0.01$, and $\lambda_r = 1$, respectively. We use smaller weights for the IMU accelerations since they are noisy. During the run-time optimization stage, we first freeze the VQ-VAE decoder $\mathcal{D}$ and then optimize the VQ-VAE latent vector $Q$ by employing the L-BFGS~\cite{liu1989limited} method with a learning rate of 1 and a convergence tolerance of $1 \times 10^{-6}$. The optimization process runs for a maximum of 1,000 iterations, maintains a history size of 200, and utilizes the strong Wolfe~\cite{nocedal2000numerical} conditions for line search.

\subsection{Training Details of Multi-Modal LLM for Motion Understanding}

During the pre-training phase, we train only the motion embedding layer $E_M$ while keeping all other modules frozen. The embedding layer is trained for 1 epoch using the Adam optimizer with a learning rate of $1 \times 10^{-3}$ and a batch size of 16.
In the multi-modal fine-tuning phase, we keep the CLIP model and multi-modal encoder frozen while fine-tuning both the image and motion embedding layers along with the large language model. We employ LoRA~\cite{hu2021lora} with a rank of 128 and an alpha value of 256. The language model is fine-tuned for 4 epochs using the Adam optimizer with a learning rate of $2 \times 10^{-5}$ and a batch size of 16.



\section{Results on Different IMU Setups}
In this section, we present results for human motion capture across different IMU configurations in Table~\ref{table:result_each_imu}. The intuitive results can be seen in~\Cref{fig:result_mocap_radar}.

From the results, we observe that human motion capture accuracy decreases when fewer IMU inputs are used. This is expected, as a lower number of IMUs provide less information, leading to greater ambiguity in the motion capture process.

Additionally, we find that using lower-body IMUs generally leads to higher motion capture accuracy compared to upper-body IMUs, a trend that is particularly evident in the Ego4o-IMU results. A possible explanation is that lower-body movements provide essential kinematic constraints for motion analysis, such as differentiating between standing and sitting postures—something a wrist-mounted IMU, for instance, cannot reliably capture. However, this pattern is less observed in the Ego4o method, likely because the text-based motion descriptions and egocentric image inputs provide contextual motion cues, enabling the model to infer postural states (e.g., standing or sitting) more effectively, thereby reducing reliance on IMU data alone.

\section{Ablation Study on Different IMU Setups}

In this section, we present ablation study results for human motion capture across different IMU configurations in Table~\ref{table:ablation_study_each_imu}.

\section{Evaluation Metrics}

We evaluate our method using three standard metrics for human motion capture accuracy: Mean Per Joint Position Error (MPJPE), Procrustes-aligned Mean Per Joint Position Error (PA-MPJPE) and Jitter. MPJPE measures the average Euclidean distance between predicted and ground truth joint positions. To compute PA-MPJPE, we first perform rigid alignment of the predicted pose to the ground truth using Procrustes analysis~\cite{kendall1989survey}, then calculate the MPJPE. The Procrustes alignment helps evaluate pose accuracy independent of global position and orientation.
Jitter~\cite{flash1985coordination} quantifies motion smoothness by measuring the mean jerk (third-time derivative of position) of all body joints in global space, expressed in $km/s^3$.

We evaluate our method with three metrics for motion description accuracy: BERT score~\cite{zhang2019bertscore}, BLEU~\cite{papineni2002bleu}, and ROUGE-L~\cite{lin2004rouge}.  BLEU measures the precision of n-gram matches between generated and reference texts, indicating how well the generated descriptions align with ground truth at the phrase level. BERT score leverages pre-trained BERT embeddings to compute semantic similarity between generated and reference descriptions, providing a more contextually-aware evaluation than traditional n-gram based metrics. ROUGE-L computes the longest common subsequence between generated and reference descriptions, capturing the fluency and sequential consistency of the generated text.

In our experiments, we employ the Python ``evaluate'' package from the Huggingface to compute BERT, BLEU, and ROUGE-L scores. For the BERT score calculation, we enable IDF weighting and rescale with baseline, setting both parameters to ``True''.

\section{w/o Part-Aware VQ-VAE}

In this section, we evaluate the effectiveness of our part-aware VQ-VAE by comparing its reconstruction accuracy with that of the traditional VQ-VAE.

Our part-aware VQ-VAE achieves a Mean Per Joint Position Error (MPJPE) of 44.93 mm and a Procrustes-aligned MPJPE (PA-MPJPE) of 32.72 mm. In contrast, the traditional VQ-VAE yields higher errors with an MPJPE of 47.73 mm and a PA-MPJPE of 36.71 mm. These results demonstrate that our part-aware approach reduces the reconstruction error, indicating superior performance in preserving motion details and overall pose structure.



\begin{figure}
    \centering
    \begin{minipage}{0.22\textwidth}
        \includegraphics[width=\textwidth]{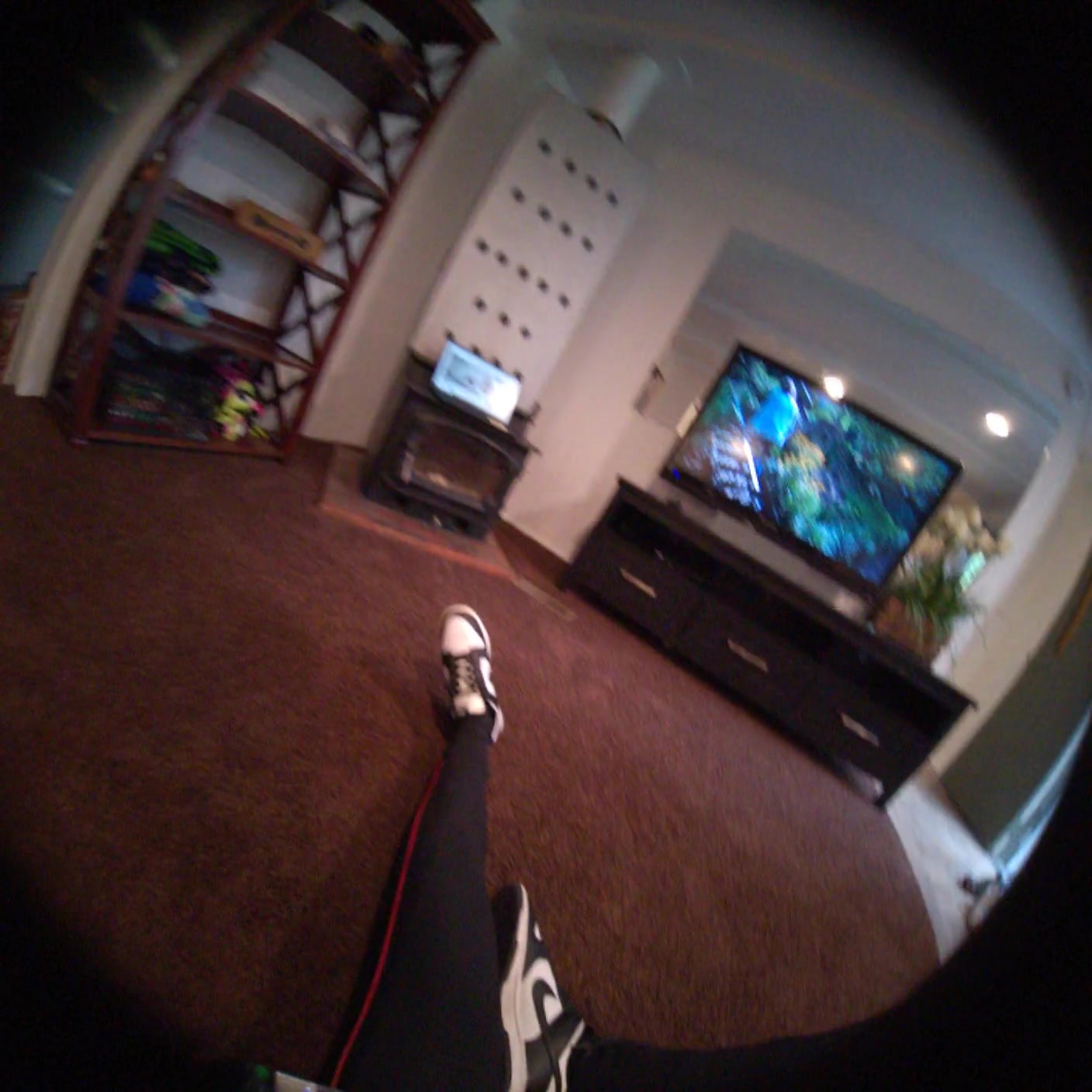}
        \label{fig:image1}
    \end{minipage}
    \hfill
    \begin{minipage}{0.22\textwidth}
        \includegraphics[width=\textwidth]{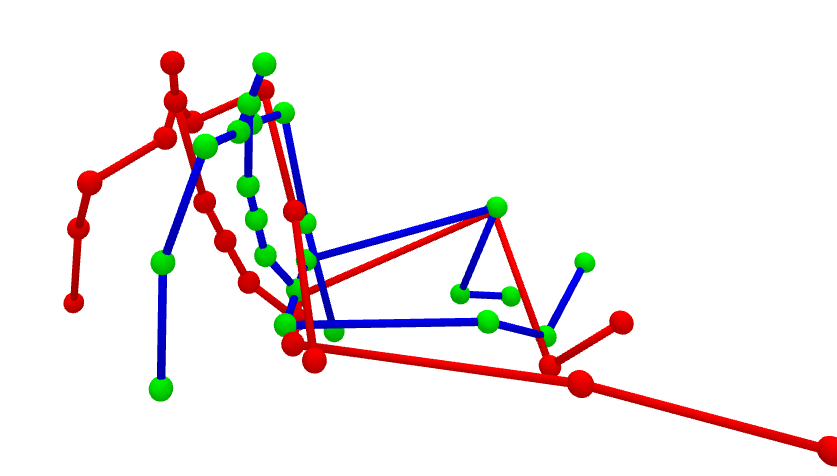}
        \label{fig:image2}
    \end{minipage}
    \caption{Failure Case. Left: input image; Right: output human body pose. Red skeleton is the ground truth pose.}
    \label{fig:failure_case}
\end{figure}

\section{Comparison with HMD-Poser}
A direct comparison between our Ego4o method and HMD-Poser~\cite{dai2024hmd} would be unfair, as Ego4o supports an arbitrary number of IMUs and multi-modal inputs (e.g., text and images), whereas HMD-Poser relies on fixed 6DoF head and hand tracking data and cannot handle multi-modal inputs. Nonetheless, we retrained HMD-Poser and evaluated it on the DIP-IMU dataset under our experimental setup. The results—87.6 mm MPJPE, 66.9 mm PA-MPJPE, and 0.12 \(km/s^3\) jitter—are inferior to ours.

\section{Failure Case}
Since IMUs track acceleration rather than position, our method may fail when the body remains still and the image lacks contextual information. This is shown in \Cref{fig:failure_case}, where the upper body and feet predictions are incorrect.

\section{Efficiency and Resource Utilization}
Our framework demonstrates real-time performance and low memory consumption across tasks:

\begin{itemize}
    \item \textbf{Motion Capture Model}
    \begin{itemize}
        \item Single image: \textbf{8.2\,ms/frame} inference speed, \textbf{0.90\,G} GPU memory
        \item 10 images: \textbf{8.6\,ms/frame} ($+4.9\%$ latency), \textbf{0.92\,G} memory ($+2.2\%$ usage)
    \end{itemize}
    
    \item \textbf{Motion Description Generation Model}
    \begin{itemize}
        \item Single image: \textbf{37.3\,ms/token} inference speed, \textbf{16.05\,G} GPU memory
        \item 10 images: \textbf{38.6\,ms/token} ($+3.5\%$ latency), \textbf{16.07\,G} memory ($+0.1\%$ usage)
    \end{itemize}
\end{itemize}

Both components show stable computational costs under increased input scales (1$\rightarrow$10 images), demonstrating minimal computational overhead when scaling to multi-image inputs. 

\end{document}